%% file: ijcai22.tex
\documentclass{article}
\usepackage{ijcai22}

\usepackage{times}
\usepackage{soul}
\usepackage{url}
\usepackage[hidelinks]{hyperref}
\usepackage[utf8]{inputenc}
\usepackage[small]{caption}
\usepackage{graphicx}
\usepackage{amsmath}
\usepackage{amsthm}
\usepackage{booktabs}
\usepackage{times}
\usepackage{latexsym}
\usepackage{comment}

\usepackage{amsfonts}
\usepackage{multicol}
\usepackage{multirow}
\usepackage{colortbl}
\usepackage{stfloats}
\usepackage[dvipsnames]{xcolor}
\usepackage{makecell}

\usepackage{algorithm}
\usepackage{algorithmic}

\usepackage{color}
\usepackage{cleveref}

\title{Neural Subgraph Explorer: Reducing Noisy Information via \\Target-Oriented Syntax Graph Pruning}

\author{Bowen Xing$^1$ \And Ivor Tsang$^{2,1}$ \\ 
\affiliations
$^1$Australian Artificial Intelligence Institute, University of Technology Sydney, Australia \\
$^2$Centre for Frontier AI Research, A*STAR, Singapore\\ 
bwxing714@gmail.com, ivor\_tsang@ihpc.a-star.edu.sg}

\begin{document}

\maketitle

\input{abstract}
\input{introduction}
\input{framework}

\input{experiment}

\input{relatedwork}

\input{conclusion}
\section*{Acknowledgements}
This work was supported by Australian Research Council  Grant (DP180100106 and DP200101328).
Ivor W. Tsang was also supported by A$^*$STAR Centre for Frontier AI Research.

\bibliographystyle{named}
\bibliography{ref}

\end{document}

%% file: abstract.tex
\begin{abstract}
Recent years have witnessed the emerging success of leveraging syntax graphs for the target sentiment classification task.
However, we discover that existing syntax-based models suffer from two issues:
\textit{noisy information aggregation} and \textit{loss of distant correlations}.
In this paper, we propose a novel model termed Neural Subgraph Explorer, which 
(1) reduces the noisy information via pruning target-irrelevant nodes on the syntax graph;
(2) introduces beneficial first-order connections between the target and its related words into the obtained graph.
Specifically, we design a multi-hop actions score estimator to evaluate the value of each word regarding the specific target.
The discrete action sequence is sampled through Gumble-Softmax and then used for both of the syntax graph and the self-attention graph.
To introduce the first-order connections between the target and its relevant words, the two pruned graphs are merged.
Finally, graph convolution is conducted on the obtained unified graph to update the hidden states.
And this process is stacked with multiple layers.
To our knowledge, this is the first attempt of target-oriented syntax graph pruning in this task.
Experimental results demonstrate the superiority of our model, which achieves new state-of-the-art performance.

\end{abstract} 

%% file: introduction.tex
\section{Introduction}
Target (or aspect) sentiment classification (TSC) \cite{TDLSTM} aims to infer the sentiment polarity of the specific target included in a review context.
For example, in the review ``The price is reasonable but the service is poor.'', there are two targets `price' and `service' with opposite sentiment polarities.
Generally, TSC task is formulated as a classification task whose input is a given context-target pair.

Earlier dominant neural TSC models are based on attention mechanisms \cite{IAN,Tencent}, which are designed to capture the correlations between the target and its relevant context words.
Although promising progress has been achieved, researchers discovered that attention mechanisms may mistakenly attend to the target's syntactically unrelated words and have difficulty capturing the distant while crucial context words \cite{asgcn,graphatt}.
To this end, the syntax graph of the context is widely leveraged to incorporate the syntactic information via applying graph neural networks (GNNs), e.g. graph convolutional network (GCN) and graph attention network (GAT).
With the combination of the BERT \cite{bert} which has proven its power in heterogeneous tasks, the recent models of the BERT+Syntax paradigm has achieved state-of-the-art performances \cite{DGEDT,dualgcn,dignet}. 
\begin{figure}[t]
 \centering
 \includegraphics[width = 0.48\textwidth]{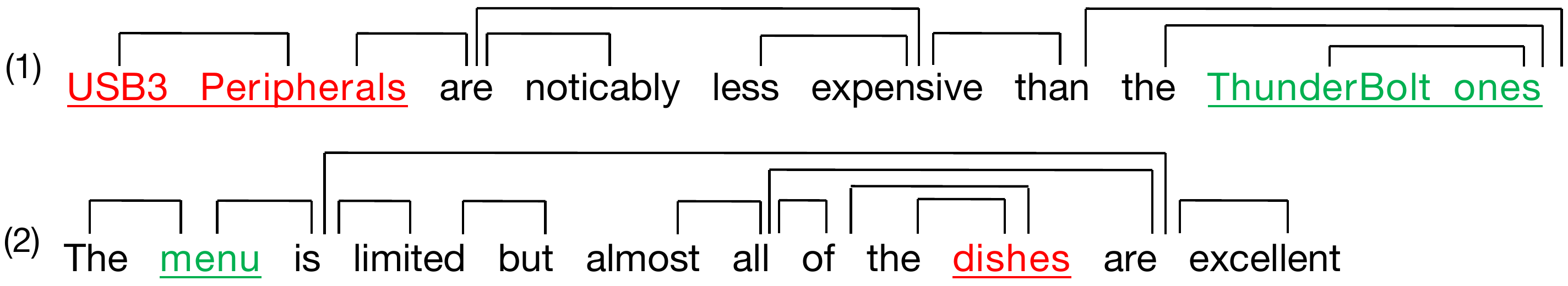}
 \caption{Illustration of three examples with syntax dependencies. Example (1) is from Laptop14 dataset while example (2) is from Restaurant14 dataset. Targets are underlined. \textcolor{Red}{Red} color denotes the sentiment of the target is positive, while \textcolor{Green}{color} denotes negative.}
 \label{fig: example}
\end{figure}

However, despite the remarkable improvements brought by the incorporation of syntactic information, we find existing models suffer from two issues which are illustrated in Fig. \ref{fig: example}:
\begin{itemize}
\item \textit{Noisy information aggregation}.
In existing models, the message passing process of GNNs is conducted on the whole syntax graph, and all words transfer information with their neighbors.
Consequently, the noisy information contained in target-irrelevant works may be aggregated into the target nodes, which disturbs the prediction.
 As shown in example (1), `less expensive' are crucial for inferring the positive sentiment of `USB3 Peripherals' and their distance on the syntax graph is 2.
However, the distance between `less expensive' and another target `ThunderBolt ones' is also 2, while this review expresses negative sentiment on `ThunderBolt ones'.
In this case, the information of `less expensive' is aggregated into both targets.
For the first target, it is beneficial, while for the later one, its information is noisy, harming the prediction. 
\item \textit{Loss of distant correlations}. 
After $l$-layer GNNs, the information of a node's $\leq l$th-order neighbors can be aggregated into it, while further nodes cannot exchange information with it.
Sometimes, the critical words may be distant from the target on the syntax graph. Thus the target loses their crucial information.
As shown in example (2), `excellent' plays the key role in expressing the positive sentiment of `dishes', while their distance is 4, which means that their correlation cannot be captured by the widely-adopted 2- or 3-layer GNNs \cite{asgcn,DGEDT,dualgcn}.
\end{itemize}

To solve the first issue, target-irrelevant nodes are supposed to be removed from the syntax graph.
For the second issue, an alternative solution is to increase the layer number of GNNs.
However, this would lead to over-fitting problem and exacerbates the first issue.
Another alternative is leveraging the fully-connected self-attention graph \cite{DGEDT,dualgcn} to introduce first-order connections.
In this case, massive correlative information between the words is introduced.
Although the beneficial first-order connections between the target and its related words are introduced, the noisy ones between the target and noisy words are also integrated, which exacerbates the first issue.
Thus we should introduce the beneficial first-order connections and eliminate the noisy first-order connections simultaneously.

In this paper, we argue that it is urgent to conduct target-oriented syntax pruning.
On the one hand, it can reduce the noisy information via pruning the target-irrelevant nodes.
On the other hand, pruning the syntax graph and self-attention graph then merging them can adaptively introduce the beneficial first-order connections between the target and its related words rather than all first-order correlations, which include the noisy ones.
To this end, we propose Neural Subgraph Explorer, whose core is a stacked target-oriented syntax graph pruning layer.
We design a multi-hop action score estimator to evaluate the contribution of each node regarding the target.
And we leverage the Gumble-Softmax \cite{gumble-softmax} for stable and differentiable discrete action sampling.
To obtain the first-order connections between the target and its related words, we apply the non-local self-attention to generate the fully-connected self-attention graph.
Then the syntax graph and self-attention graph are pruned and merged into a unified graph, on which the proposed position weighted GCN is applied for message passing.
Another advantage of introducing the self-attention graph is that it guarantees the connectivity of the obtained graph, resolving the potential issue that the pruned syntax graph has isolated nodes which harms the message passing.  
Experimental results on benchmark datasets show that our model achieves new state-of-the-art performance, significantly surpassing existing models.

%% file: framework.tex

\section{Methodology}\label{sec: framework}
\begin{figure}[t]
 \centering
 \includegraphics[width = 0.48\textwidth]{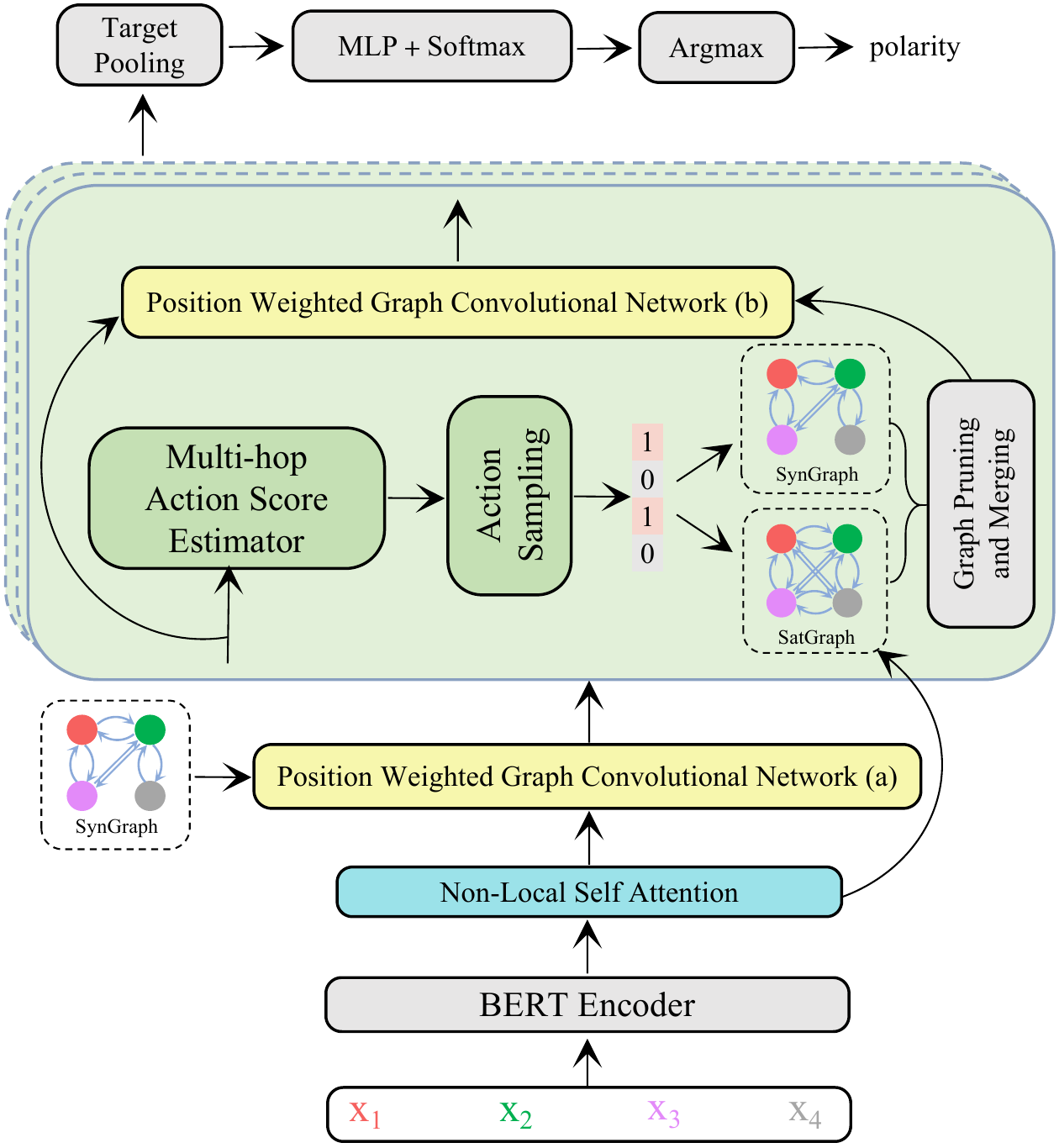}
 \caption{The architecture of Neural Subgraph Explorer. SynGraph is obtained from off-the-shelf dependency parser.}
 \label{fig: model}
\end{figure}

The architecture of our model is illustrated in Fig. \ref{fig: model}.
We propose the position weighted GCN for graph message passing.
And the core of our model is the proposed target-oriented syntax-graph pruning module.
Next, we introduce the details.

\subsection{Contextual and Syntactic Encoding} \label{sec: encoding}
\paragraph{BERT Encoder.}
Following the up-to-date models \cite{RGAT,tgcn}, we employ BERT encoder to obtain the initial word hidden states.
Given the review context word sequence ${x_1, x_2, ..., x_{N_c}}$ and the target word sequence ${a_1, ..., a_{N_t}}$, the input of BERT is the target-context pair:
\begin{equation}
 \small \langle[\text{CLS}]; {x_1, x_2, ..., x_{N_c}}; \text{[SEP]}; {a_1, ..., a_{N_t}}; [\text{SEP}]\rangle
\end{equation}
 where $N_c$ and $N_t$ are the length of review context and target respectively, and $\langle;\rangle$ denotes sequence concatenation.
Then we obtain the hidden state sequence of the review: $\hat{H_c}=[h^c_1, ..., h^c_{N_c}]\in\mathbb{R}^{N_c \times d}$, including the target word hidden states $\hat{H_t} = [h^t_1, ..., h^t_{N_t}]\in\mathbb{R}^{N_t \times d}$, where $d$ denotes the dimension of the hidden state.
\paragraph{Position Weighted Graph Convolutional Network.}
Graph convolutional network (GCN) has been widely used to encode the syntax graph to integrate syntactic information into hidden states.
In this work, based on standard GCN, we introduce a weight for each word to indicate its position relative to the target, forming the position-weighted GCN.
By this means, the potential relative words for the target can be highlighted.
Specifically, the node updating process can be formulated as:
\begin{equation}
\small
\begin{aligned}
{h}_{i}^{l}=& \operatorname{ReLU}\left(\sum_{j=0}^N \frac{A^{syn}_{ij}(w_p^j{W_g^l} h_j^{l-1})}{d_i + 1} + b_g^l\right)\\
  \mu_j=& \frac{ j-\tau}{N+1}\\
   \ w_p^j =& 1 - \vert\mu_j\vert
\end{aligned}
\end{equation}
where $A^{syn}$ is the adjacent matrix derived from the dependency parsing result and $A^0_{ij} = 1$ if there is a dependency from node $j$ to $i$; $\mu_j$ is the relative offset between $j$-th word and the target and $w_p^j$ is the position weight of $j$-th word; $d_i$ denotes the degree of $i-$th node; $W_g^l$ and $b_g^l$ are parameters.
If the target is a phrase, $t-\tau$ is calculated with its left or right boundary index according to which side the word locates.

Now we obtain the syntax-enhanced context hidden states $H_c$ and target hidden states $H_t$.
And by applying mean pooling over $H_t$, we obtain the initial target representation $r_t$.

\subsection{Target-oriented Syntax Graph Pruning}
The process of target-oriented syntax graph pruning includes four steps:
(1) the multi-hop action score estimator evaluates the value of each word regarding the target, producing the degree that whether the word should be pruned or reserved;
(2) the Gumble-Softmax is leveraged for differentiable discrete action sampling and generate the action sequence;
(3) the action sequence is used to mask the adjacent matrix of the syntax graph and self-attention graph;
(4) the position weighted GCN is adopted for message passing on the obtained graph, updating the hidden states.
Next we present the details of each module following the above order.
\subsubsection{Multi-hop Action Score Estimator}
To decide whether a node on the graph should be pruned, each word is supposed to be assigned a score to represent its contribution for expressing the sentiment semantics of the given target.
To this end, inspired from \cite{Tencent}, we design a multi-hop action score estimator.
At each hop, it produces a gate score for each word and a summarized target-centric context vector which is used at next hop. 
The details are given bellow:
\begin{equation}
\small
\begin{aligned}
s_{j}^{t,t'}&=W_{s}\left[w_p^j h^{t,t'}_j, \mu_j, C_{t-1}, r_t\right]+b_{s}\\
g^{t,t'}_j &= \operatorname{Sigmoid}(s_j^{t,t'})\\
\alpha_{j}^{t}&=\frac{\exp \left(s_{j}^{t,t'}\right)}{\sum_{k}^{N_c} \exp \left(s_{k}^{t,t'}\right)}\\
{I}^{t,t'}&=\sum_{j}^{N_c} \alpha_{j}^{t} w_j h^{t,t'}_j \\
C_{t} &= \operatorname{GRU}(I^{t,t'}, C_{t-1})
\end{aligned}
\end{equation}
where $W_s$ and $b_s$ are parameters; $\operatorname{GRU}$ denotes gated recurrent unit.
$C_0$ is initialized as a zero vector.

Then the output gate $g^{t,T'}_j$ of final step $T'$ can derive 
\begin{equation}
\begin{aligned}
p^t_{j,0}&=1-g^{t,T'}, \
p^t_{j,1}&=g^{t,T'}_j
\end{aligned}
\end{equation}
 which represent the possibilities that node $j$ should be pruned or reserved, respectively.

\subsubsection{Action Sampling}
Now we have a problem of discrete action selecting.
Although REINFFORCE algorithm \cite{Williams_1992} are commonly used for this problem, it causes model instability and hard training.
To this end, we leverage the Gumbel-Softmax \cite{gumble-softmax} trick for differentiable action sampling:
\begin{equation}
\small act_{j}^{t}=\frac{\exp \left(\left(\log \left(p_{j, 1}^{t}\right)+\epsilon_{1}\right) / \pi\right)}{\left.\sum_{a=0}^{1} \exp \left(\left(\log \left(p_{i, a}^{t}\right)+\epsilon_{t}\right) / \pi\right)\right)}
\end{equation}
where $\epsilon_t$ is randomly sampled from Gumbel distribution
and $\pi$ is the temperature coefficient which is set 0.1 in this work.
$act_{j}^{t}=0$ denotes pruning while $act_{j}^{t}=1$ denotes reserving. 

\subsubsection{Graph Pruning and Merging}
Given the set of actions $\{act_j^t\}_{i=1}^{N_c}$ corresponding to the context word, we use it to mask $A^{syn}$:
 \begin{equation}
 A^{syn,t}_{ij}=act_j^t\cdot A^{syn}_{ij}
 \end{equation}
  where $A^{syn,t}$ is the adjacent matrix of the pruned syntax graph of the $t$-th layer.

Now in the pruned syntax graph, some target-irrelevant nodes (words) are removed.
However, the issue of \textit{loss of distant correlations} is not tackled.
Besides, the pruning operation on the whole syntax graph may lead to a potential problem that the pruned syntax graph may include isolated nodes, which hinders the message passing on the obtained graph.

To solve the above two issues, we propose to exploit the self-attention graph (SatGraph), which is a fully-connected semantic graph consisting of all words in the sentence and derived by the self-attention \cite{transformer}.
SatGraph can provide the first-order connections between each two nodes. 
Therefore, merging the pruned SatGraph with the pruned SynGraph can not only guarantee the connectivity but also can directly connect the target with its related words, promoting the aggregation of target-related information, which is beneficial for prediction.

An intuitive way to obtain the self-attention graph is to retrieve the self-attention matrix at the last layer of BERT \cite{bert}.
However, considering the self-attentions in BERT$_{\text{BASE}}$ (BERT$_{\text{LARGE}}$) are 12-head (16-head), and the word representations are segmented into 12 (16) \textit{local} subspaces, one of the 12 (16) self-attention matrices can only reflect the \textit{local} word correlations in the \textit{local} subspace rather than general global semantic space in which the subsequent modules work.
Therefore, due to the segmentation bias between the multi-head \textit{local} subspaces and the general \textit{global} semantic space, the self-attention matrix derived by BERT cannot be used for the self-attention graph.

To this end, we add a non-local self-attention layer between BERT and position weighted GCN, as shown in Fig. \ref{fig: model}, to obtain the \textit{global} self-attention matrix $A^{sat}\in\mathbb{R}^{N_c \times N_c}$ representing the correlations between words in the general semantics space which is consistent with subsequent modules.
Specifically, $A^{sat}$ is obtained as follows:
\begin{equation}
\small {A^{sat}} = \operatorname{Softmax}\left((\hat{H_c} M_q) ({\hat{H_c} M_k })^{\mathsf{T}}/\sqrt{d}\right) 
\end{equation}
Then the updated hidden states are obtained by:
\begin{equation}
 \small \widetilde{H_c}=A^{sat} \hat{H_c} M_v
\end{equation}
where $M_q, M_k, M_v \in \mathbb{R}^{d\times d}$ are parameters.
And then $\widetilde{H_c}$ is fed to the position weighted GCN before the target-oriented syntax graph pruning layer.


Then we prune SatGraph in the same way as SynGraph and merge them:
\begin{equation}
\small
\begin{aligned}
 A^{sat,t}_{ij}&=act_j^t\cdot A^{sat}_{ij} \\
A^t &= (A^{syn,t} + A^{sat})/2
\end{aligned}
\end{equation}
In this process, not only the beneficial first-order connections are introduced, but also the noisy first-order connections are removed duo to pruning operation on SatGraph.
Then $A^t$ will be used for messaging passing later.

\subsubsection{Node Updating}
For message passing on the obtained graph, we apply another position weighted GCN (noted as (b) in Fig. \ref{fig: model}) over $A^t$.
\subsubsection{Multi-layer Stacking}
In order to let our model gradually improve the graph pruning and learn deep features, we stack the target-oriented syntax graph pruning module in a multi-layer manner.
\subsection{Prediction and Training}
After $T$ layers of target-oriented syntax graph pruning, we obtain the final representations of each context word, which includes $N_t$ target words.
We apply mean pooling over all target words to generate the final target representation, getting the final target representation $R_t$

Then we fed $R_t$ to a multi-layer perception (MLP) and then $softmax$ for classification:
\begin{equation}
 \small P=\operatorname{softmax} \left(W_c^1 (W_c^2 R_{a}+b_c^2) + b_c^1\right)
\end{equation}
Finally, by applying $arg\ max$ on the class vector $P$, we can get the produced sentiment polarity of the target in the review.

Given $D$ training samples, the training objective is:
\begin{equation}
\small \ell=-\sum_{i=1}^{D} \sum_{c \in \mathcal{C}} I(y=c) \log (P(y=c))
\end{equation}
where $y$ is the ground-truth class, $I$ is an indicator function, and $\mathcal{C}$ denotes the sentiment polarity class set.

%% file: experiment.tex
\section{Experiment}\label{sec:experiment}
\subsection{Experimental Setup}
\subsubsection{Dataset}
We conduct experiments on three public benchmark datasets to obtain reliable and authoritative results.
Restaurant14 and Laptop14 are from \cite{Semeval2014}, and Restaurant15 is from \cite{semeval2015}.
We pre-process the datasets following the same way as previous works \cite{asgcn,RGAT,tgcn}.
The statistics of all datasets are shown in Table \ref{table: dataset}.

\begin{table}[ht]
\centering
\fontsize{9}{11}\selectfont
\setlength{\tabcolsep}{1.5mm}{
\begin{tabular}{ccccccc}
\toprule
\multirow{2}{*}{Dataset} & \multicolumn{2}{c}{Positive} & \multicolumn{2}{c}{Neutral} & \multicolumn{2}{c}{Negative} \\\specialrule{0em}{0pt}{1pt} \cline{2-3} \cline{4-5} \cline{6-7}\specialrule{0em}{1pt}{1.5pt}
                         & Train         & Test         & Train         & Test        & Train         & Test         \\ \specialrule{0em}{0pt}{1pt}\hline \specialrule{0em}{1.5pt}{1.5pt}
Laptop14                    & 994           & 341          &  464          &    169      &  870          &    128       \\
Restaurant14                    & 2164          & 728          & 637           & 196         & 807           & 196          \\
Restaurant15                    & 912           & 326          & 36            & 34          & 256           & 182         \\\bottomrule
\end{tabular}}
\caption{Dataset statistics of the three datasets.}
\label{table: dataset}
\end{table}

\subsubsection{Implementation Details}
We adopt the BERT$_{\text{BASE}}$ uncased version as the BERT encoder and it is fine-tuned in the experiments. 
We train our Neural Subgraph Hunter using AdamW optimizer.
The dependency parser used in our experiments is from spaCy toolkit\footnote{https://spacy.io/.}.
In our experiments, the dimension of hidden units are 768.
The dropout rate for BERT encoder is 0.1, while the dropout rate for other modules is 0.3.
The batch size is 16 and epoch number is 30.
The learning rates are 1e-5, 5e-5, 3e-5 for Lap14, Res14 and Res15 datasets respectively.
The weight decay rages are $0.05$ for Res14 and Res15 datasets while $0.001$ for Lap14 datasets.
The layer number of the position weighted GCN (a) and (b)are both 2.
The hop number of the multi-hop action score estimator is 3.
The layer number of target-oriented syntax graph pruning is 2.
And our source code is available at \url{https://github.com/XingBowen714/Neural-Subgraph-Explorer}.

Accuracy (Acc) and Macro-F1 (F1) are used as evaluation metrics.
Since there is no official validation set for the datasets, we report the average results over three random runs.

\subsubsection{Baselines for Comparison}
The baselines can be divided into four categories regarding whether BERT and syntax are leveraged:\\
(A) BERT $\times$ Syntax $\times$:
1. IAN \cite{IAN} separately encodes the target and context, then models their interactions through an interactive attention mechanism.
2. RAM \cite{Tencent} uses a GRU attention mechanism to recurrently extracts the target-related semantics.\\
(B) BERT $\times$ Syntax $\checkmark$:
3. ASGCN \cite{asgcn} utilizes GCN to leverage syntactic information.
4. BiGCN \cite{bigcn} employs GCN to convolute over hierarchical syntactic and lexical graphs.\\
(C) BERT $\checkmark$ Syntax $\times$:
5. BERT-SPC \cite{bert} takes the concatenated context-target pair as input and uses the output hidden state of [CLS] token for classification.
6. AEN-BERT \cite{aen-bert} employs multiple attention layers to learn target-context interactions.\\
(D) BERT $\checkmark$ Syntax $\checkmark$:
7. ASGCN+BERT \cite{asgcn}. Since the backbone of our Neural Subgraph Explore is BERT+GCN, we augment the ASGCN model with BERT encoder to form a baseline.
8. KGCapsAN-BERT \cite{kgcap} utilizes multi-prior knowledge to guide the capsule attention process and use a GCN-based syntactic layer to integrate the syntactic knowledge. 
9. R-GAT+BERT \cite{RGAT} uses the relational graph attention network to aggregate the global relational information from all context words into the target node representation. 
10. DGEDT-BERT \cite{DGEDT} employs a dual-transformer network to model the interactions between the flat textual knowledge and dependency graph empowered knowledge.  
11. A-KVMN+BERT \cite{kvmn-eacl} uses a key-value memory network to leverage not only word-word relations but also their dependency types.
12. BERT+T-GCN \cite{tgcn} leverages the dependency types in T-GCN and uses an attentive layer ensemble to learn the comprehensive representation from different T-GCN layers.
13. DualGCN+BERT \cite{dualgcn} uses orthogonal and differential regularizers to model the interactions between semantics and syntax.

Note that all of the BERT encoders in baselines are BERT-base uncased version, the same as ours.
And for fair comparison, we reproduce the average results of R-GAT+BERT, A-KVMN+BERT, BERT+T-GCN and DualGCN+BERT on three random runs because they report the best results rather than average results in their original paper. 
\subsection{Main Results}
\begin{table*}[ht]
\fontsize{9}{12}\selectfont
\centering
\setlength{\tabcolsep}{3mm}{
\begin{tabular}{c|cccccc}\toprule
 \multirow{2}{*}{Model} & \multicolumn{2}{c}{Laptop14} & \multicolumn{2}{c}{Restaurant14} & \multicolumn{2}{c}{Restaurant15}  \\\cline{2-7} 
      &Acc (\%)     &F1 (\%)      &Acc  (\%)     &F1  (\%)      &Acc (\%)        &F1 (\%)      \\\midrule
IAN$^\ddag$ \cite{IAN} &72.05     & 67.38    &79.26    &70.09  & 78.54    & 52.65\\
RAM$^\ddag$ \cite{Tencent} &74.49& 71.35 & 80.23 &70.80&79.30&60.49\\ \hline
 ASGCN \cite{asgcn} & 75.55    &71.05     &80.77    & 72.02  & 79.89   & 61.89\\
 BiGCN  \cite{bigcn} & 74.59    &71.84    &81.97     &73.48    & 81.16    & 64.79  \\ \hline
 BERT-SPC \cite{bert} & 78.47    &73.67  &84.94   & 78.00   & 83.40    &  65.00   \\
 AEN-BERT \cite{aen-bert}  & 79.93    &76.31  &83.12   & 73.76   & -    &  -   \\ \hline
 ASGCN+BERT$^\dag$ \cite{asgcn}    & 78.92    &74.35     &85.87    & 79.32  & 83.85   & 68.73        \\
KGCapsAN-BERT \cite{kgcap}  & 79.47    &76.61    &85.36      & 79.00   & - & - \\
R-GAT+BERT$^\dag$ \cite{RGAT}& 79.31    &75.40  &86.10   & 80.04   & 83.95    &  69.47   \\
DGEDT-BERT \cite{DGEDT}  & 79.8    &75.6    &86.3      & 80.0   & 84.0   &71.0    \\
 A-KVMN+BERT$^\dag$ \cite{kvmn-eacl}  & 79.20    &75.76   &85.89      & 78.29   & 83.89  &67.88    \\
 BERT+T-GCN$^\dag$  \cite{tgcn}  & 80.56    &76.95    &85.95      & 79.40   & \underline{84.81} & 71.09 \\
 DualGCN+BERT$^\dag$  \cite{dualgcn}  & \underline{80.83} & \underline{77.35}   & \underline{86.64}  &  \underline{80.76} &84.69&\underline{71.58}\\
\midrule
Neural Subgraph Explorer (ours) & \textbf{82.13}   &\textbf{78.51}     &\textbf{87.35}       & \textbf{82.04}    &\textbf{86.29} &\textbf{74.43} \\
\bottomrule
\end{tabular}}
\caption{Performances comparison (in \%). $^\dag$ indicates we reproduce the results using the official source code, and $^\ddag$ denotes that the results are retrieved from [Zhang \textit{et al}., 2019].
Our Neural Subgraph Explorer outperforms previous SOTA models and corresponding baselines on all datasets, being statistically significant ($p < 0.05$ under t-test.)}
\label{table: results}
\end{table*}
The performances of our Neural Subgraph Explorer and baselines are shown in Table \ref{table: results}.
We can observe that BERT can significantly boost the performance while integrating syntactic information into BERT-based models can bring further significant improvement.
And the best-performing models are all based on BERT+Syntax paradigm.

However, state-of-the-art models neglect the problem that not all words are useful for expressing the sentiment of the specific target, and they just adopt GNNs to conduct message passing on the whole syntax graph.
As a result, the noisy information from the target-irrelevant words is aggregated into the target's and its related words' node representation, which affects the prediction of the specific target's sentiment.
To overcome this problem, we propose a Neural Subgraph Explorer model which can adaptively and dynamically prune the noisy nodes corresponding to the target-irrelevant words.
As shown in Table \ref{table: results}, our model achieves new state-of-the-art performance, obtaining consistent improvements over all baselines in terms of both Acc and F1.
Specifically, our model surpasses previous best scores by 1.30\%, 0.71\%, and 1.16\% in terms of Acc on Lap14, Res14, and Res15 datasets, respectively.
And in terms of F1, our model surpasses previous best scores by 1.16\%, 1.28\%, and 3.44\% on Lap14, Res14, and Res15 datasets, respectively.
All the improvements are attributed to the discarding of noisy information via pruning noisy nodes on the vanilla syntax graph and introducing first-order dependency to facilitate the aggregation of distant while crucial target-related information.

\subsection{Ablation Study}
\begin{table}[ht]
\centering
\fontsize{9}{12}\selectfont
\setlength{\tabcolsep}{2mm}{
\begin{tabular}{ccccccc}
\toprule
\multirow{2}{*}{Variants} & \multicolumn{2}{c}{Laptop14} & \multicolumn{2}{c}{Restaurant14} & \multicolumn{2}{c}{Restaurant15}  \\\cline{2-7} 
      &Acc      &F1      &Acc       &F1       &Acc       &F1       \\\midrule
Full Model  &\textbf{82.13}   &\textbf{78.51}     &\textbf{87.35}       & \textbf{82.04}    &\textbf{86.29} &\textbf{74.43} \\ \midrule
\multirow{2}{*}{NoPrune}      &79.52 &75.70    &86.04     &79.77     &84.50    &70.83 \\
                              &$\downarrow$2.61   & $\downarrow$2.81   &  $\downarrow$1.31   & $\downarrow$2.27    &$\downarrow$1.78    & $\downarrow$3.60\\ \hline
\multirow{2}{*}{RandPrune}    &80.67  &76.97 &86.49   &80.44  &84.75   &72.10   \\
                              &$\downarrow$1.46  & $\downarrow$1.54  &$\downarrow$ 0.86  & $\downarrow$1.60 &  $\downarrow$1.54 & $\downarrow$2.33  \\ \hline
\multirow{2}{*}{NoMerge}    &81.61  &78.25 &87.05   &81.64  &85.79   &73.29   \\
                              &$\downarrow$0.52  & $\downarrow$0.26  &$\downarrow$ 0.30  & $\downarrow$0.40 &  $\downarrow$0.50 & $\downarrow$1.14  \\
\bottomrule
\end{tabular}}
\caption{Results of ablation experiments.}
\label{table: ablat}
\end{table}
We conduct ablation experiments to look into Neural Subgraph Explorer and investigate how it works well.
The experiment results are shown in Table \ref{table: ablat}.

We first study the effect of target-oriented syntax graph pruning via removing the pruning operation.
In practice, we achieve this by setting all $act^t_i$ as 1 and this variant is termed NoPrune.
From Table \ref{table: ablat} we can observe that NoPrune obtains much poorer results compared with the full model.
The reason is that without noisy node pruning, the main advantage of Neural Subgraph Explorer is lost and the noisy information conveyed in the target-irrelevant nodes harms the prediction.

Then we study the effect of multi-hop action score estimator, which determines whether a word should be pruned.
We design a variant named RandPrune, which assigns a random value to each $g^{t,T'}_i$.
Therefore, the nodes in the syntax graphs are randomly pruned.
From Table \ref{table: ablat} we can find that RandPrune is significantly inferior to the full model.
This is because without the multi-hop action score estimator, RandPrune cannot identify the crucial words that contribute to the expression of the target's sentiment, and the words it prunes may be the critical words and it may reserve the noisy words that harm the prediction.

Finally, we investigate the effect of graph merging via designing a variant termed NoMerge, in which the pruned SynGraph is directly used for message passing. 
From Table \ref{table: ablat} we can find that NoMerge has a considerable results drop compared with the full model.
This proves the effectiveness of merging the pruned SynGraph and pruned SatGraph, which can introduce first-order connections to facilitate the information aggregation of distant target-related words and guarantee the connectivity of the obtained graph.
\subsection{Investigation of Layer Number}
\subsubsection{\#Target-oriented Syntax Graph Pruning}
\begin{figure}[t]
 \centering
 \includegraphics[width = 0.48\textwidth]{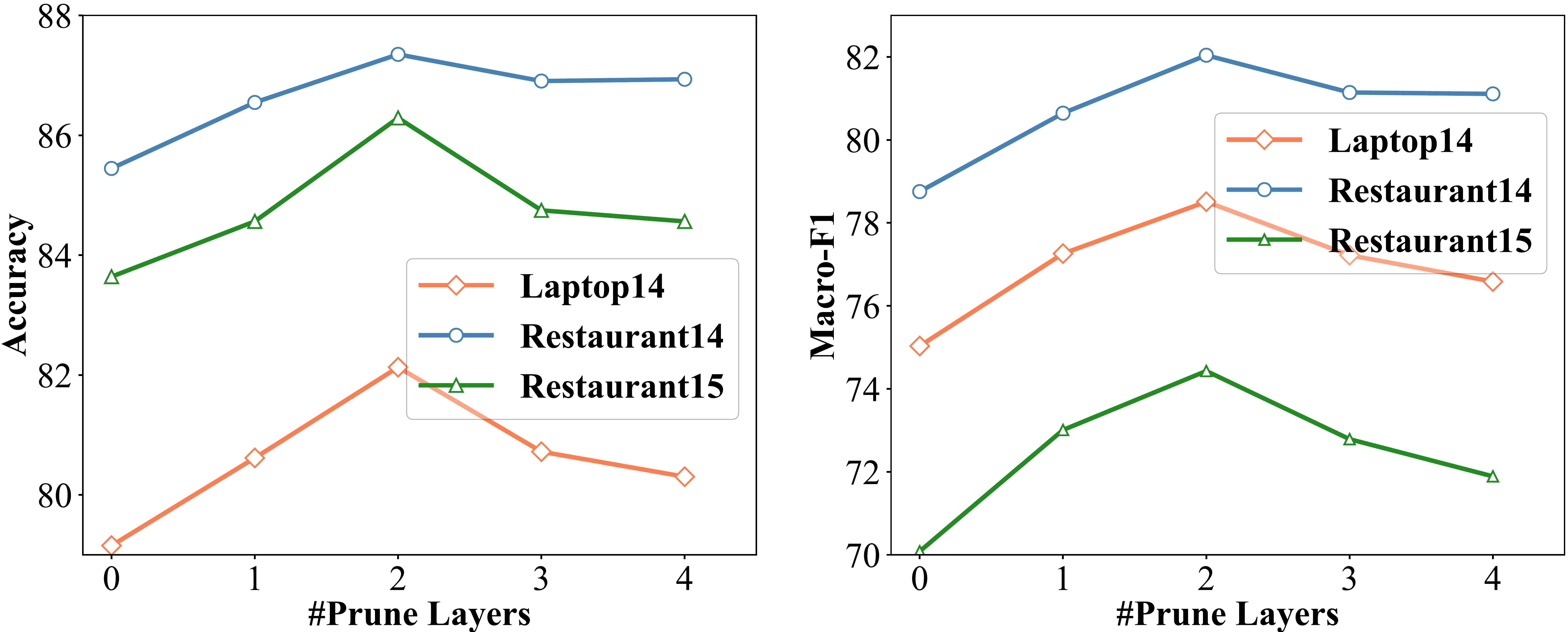}
 \caption{The impact of the number of pruning layers.}
 \label{fig: prune}
\end{figure}
To investigate the impact of layer number of target-oriented syntax graph pruning, we vary the value of $T$ from 0 to 4 and illustrate the corresponding Accuracy and Macro-F1 in Fig. \ref{fig: prune}.
$T=0$ denotes there is no target-oriented syntax graph pruning. As a result, the worst performances are obtained.
With $T$ increasing from 0 to 2, the performances increase then reach the peaks.
However, continuously increasing $T$ results in the drop of results.
There are two reasons.
The first one is that too large $T$ makes it more difficult to train the model due to a large number of parameters.
The other one is that since there is a linear relationship between $T$ and the total GCN layer number of Neural Subgraph Explorer, too large $T$ causes the over-smoothing problem due to too much node aggregation, losing the specific and important features.
\subsubsection{\#Position weighted GCN}
\begin{figure}[t]
 \centering
 \includegraphics[width = 0.48\textwidth]{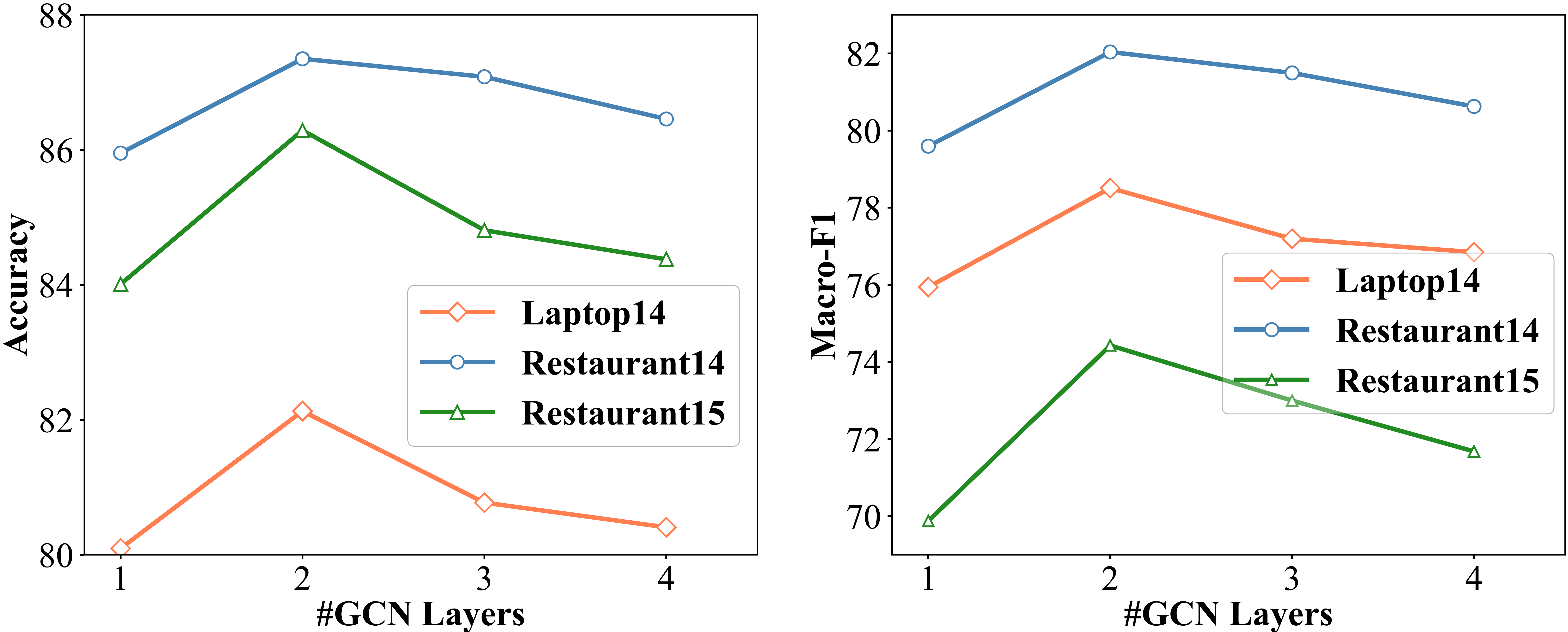}
 \caption{The impact of layer number of position weighted GCN.}
 \label{fig: gcn-layer}
\end{figure}
For each position weighted GCN in our model, the layer number $T'$ denotes that after the $T'$-layer GCN, the information of node $i$ can be aggregated into node $j$ if node $i$ is node $j$'s $t'$th-order neighbor and $t'<T'$.
When $T'=1$, only a node's directly adjacent nodes can transfer information with it, which is intuitively insufficient for capturing the sentiment features for the target.
And as shown in Fig. \ref{fig: gcn-layer}, $T'=1$ results in the worst performances.
However, too large $T'$ also has a negative impact on performances because too many GCN layers bring the over-fitting and over-smoothing problem.
Finally, as illustrated in Fig. \ref{fig: gcn-layer}, the 2-layer position weighted GCN obtains the best results for our Neural Subgraph Explorer. 

%% file: relatedwork.tex
\section{Related Works} \label{sec: related work}
In recent years, dominant TSC models are all based on neural networks which can automatically learn representations.
And attention mechanisms \cite{ATAE,IAN,Tencent,songle2019,AA-LSTM} are widely utilized to capture the target-related words.
\cite{IAN} propose an interactive attention mechanism to model the mutual interactions between target and context.
\cite{Tencent} propose a GRU-based attention mechanism to aggregate the target-centric semantics.

Recently, researchers discovered that only relying on attention mechanism is insufficient for TSC, especially for the cases in which the target is distant to its related words.
To this end, a bunch of GNN-based models \cite{asgcn,DGEDT,sagat,tgcn,dualgcn,jair,RGAT,dignet,kagrmn} are designed to encode the syntactic information for capturing the dependencies between the target and the crucial words.
\cite{sagat} propose a GAT-based model to leverage the syntactic information via applying GAT on the syntax graph, which is obtained from an off-the-shelf dependency parser.
\cite{dualgcn} propose a dual-GCN architecture to model the semantic and syntactic information. Besides, the orthogonal regularizer and the differential regularizer are proposed to learn deeper correlations between the words.
And with the power of BERT, the models based on BERT+Syntax paradigm have achieved state-of-the-art performances.

However, since existing models apply GNNs on the whole syntax graph, they suffer from two problems: \textit{noisy information aggregation} and \textit{loss of distant correlations}.
In this paper, we propose a novel model termed Neural Subgraph Explorer to solve these two problems. 

%% file: conclusion.tex
\section{Conclusion}
This paper proposes a novel Neural Subgraph Explorer model to tackle the target sentiment classification task.
On the one hand, it can discard the noisy information contained in the noisy words regarding the given target through the stacked target-oriented syntax graph pruning module. 
On the other hand, it introduces first-order connections between the target and the crucial words via merging the pruned self-attention graph with the pruned syntax graph.
In this way, more useless information can be removed and more crucial information can be captured.
Finally, experiments are conducted on benchmark datasets and the effectiveness of our model has been proven.

%% file: ijcai22.bbl
\begin{thebibliography}{}

\bibitem[\protect\citeauthoryear{Chen \bgroup \em et al.\egroup
  }{2017}]{Tencent}
Peng Chen, Zhongqian Sun, Lidong Bing, and Wei Yang.
\newblock Recurrent attention network on memory for aspect sentiment analysis.
\newblock In {\em EMNLP}, pages 452--461, 2017.

\bibitem[\protect\citeauthoryear{Devlin \bgroup \em et al.\egroup
  }{2019}]{bert}
Jacob Devlin, Ming-Wei Chang, Kenton Lee, and Kristina Toutanova.
\newblock {BERT}: Pre-training of deep bidirectional transformers for language
  understanding.
\newblock In {\em NAACL}, pages 4171--4186, 2019.

\bibitem[\protect\citeauthoryear{Huang and Carley}{2019}]{graphatt}
Binxuan Huang and Kathleen Carley.
\newblock Syntax-aware aspect level sentiment classification with graph
  attention networks.
\newblock In {\em EMNLP}, pages 5469--5477, Hong Kong, China, 2019.

\bibitem[\protect\citeauthoryear{Huang \bgroup \em et al.\egroup
  }{2020}]{sagat}
Lianzhe Huang, Xin Sun, Sujian Li, Linhao Zhang, and Houfeng Wang.
\newblock Syntax-aware graph attention network for aspect-level sentiment
  classification.
\newblock In {\em COLING}, pages 799--810, 2020.

\bibitem[\protect\citeauthoryear{Jang \bgroup \em et al.\egroup
  }{2017}]{gumble-softmax}
Eric Jang, Shixiang Gu, and Ben Poole.
\newblock Categorical reparameterization with gumbel-softmax.
\newblock In {\em ICLR (Poster)}, 2017.

\bibitem[\protect\citeauthoryear{Li \bgroup \em et al.\egroup }{2021}]{dualgcn}
Ruifan Li, Hao Chen, Fangxiang Feng, Zhanyu Ma, Xiaojie Wang, and Eduard Hovy.
\newblock Dual graph convolutional networks for aspect-based sentiment
  analysis.
\newblock In {\em ACL}, pages 6319--6329, 2021.

\bibitem[\protect\citeauthoryear{Ma \bgroup \em et al.\egroup }{2017}]{IAN}
Dehong Ma, Sujian Li, Xiaodong Zhang, and Houfeng Wang.
\newblock Interactive attention networks for aspect-level sentiment
  classification.
\newblock In {\em {IJCAI}}, pages 4068--4074, 2017.

\bibitem[\protect\citeauthoryear{Pontiki \bgroup \em et al.\egroup
  }{2014}]{Semeval2014}
Maria Pontiki, Dimitris Galanis, John Pavlopoulos, Harris Papageorgiou, Ion
  Androutsopoulos, and Suresh Manandhar.
\newblock {S}em{E}val-2014 task 4: Aspect based sentiment analysis.
\newblock In {\em {S}em{E}val 2014}, pages 27--35, 2014.

\bibitem[\protect\citeauthoryear{Pontiki \bgroup \em et al.\egroup
  }{2015}]{semeval2015}
Maria Pontiki, Dimitris Galanis, Haris Papageorgiou, Suresh Manandhar, and Ion
  Androutsopoulos.
\newblock {S}em{E}val-2015 task 12: Aspect based sentiment analysis.
\newblock In {\em {S}em{E}val 2015}, pages 486--495, 2015.

\bibitem[\protect\citeauthoryear{Song \bgroup \em et al.\egroup
  }{2019}]{aen-bert}
Youwei Song, Jiahai Wang, Tao Jiang, Zhiyue Liu, and Yanghui Rao.
\newblock Attentional encoder network for targeted sentiment classification.
\newblock {\em arXiv preprint arXiv:1902.09314}, 2019.

\bibitem[\protect\citeauthoryear{Tang \bgroup \em et al.\egroup
  }{2016}]{TDLSTM}
Duyu Tang, Bing Qin, Xiaocheng Feng, and Ting Liu.
\newblock Effective {LSTM}s for target-dependent sentiment classification.
\newblock In {\em {COLING}}, pages 3298--3307, 2016.

\bibitem[\protect\citeauthoryear{Tang \bgroup \em et al.\egroup
  }{2019}]{songle2019}
Jialong Tang, Ziyao Lu, Jinsong Su, Yubin Ge, Linfeng Song, Le~Sun, and Jiebo
  Luo.
\newblock Progressive self-supervised attention learning for aspect-level
  sentiment analysis.
\newblock In {\em ACL}, pages 557--566, 2019.

\bibitem[\protect\citeauthoryear{Tang \bgroup \em et al.\egroup }{2020}]{DGEDT}
Hao Tang, Donghong Ji, Chenliang Li, and Qiji Zhou.
\newblock Dependency graph enhanced dual-transformer structure for aspect-based
  sentiment classification.
\newblock In {\em ACL}, pages 6578--6588, 2020.

\bibitem[\protect\citeauthoryear{Tian \bgroup \em et al.\egroup }{2021a}]{tgcn}
Yuanhe Tian, Guimin Chen, and Yan Song.
\newblock Aspect-based sentiment analysis with type-aware graph convolutional
  networks and layer ensemble.
\newblock In {\em NAACL}, pages 2910--2922, 2021.

\bibitem[\protect\citeauthoryear{Tian \bgroup \em et al.\egroup
  }{2021b}]{kvmn-eacl}
Yuanhe Tian, Guimin Chen, and Yan Song.
\newblock Enhancing aspect-level sentiment analysis with word dependencies.
\newblock In {\em EACL}, pages 3726--3739, 2021.

\bibitem[\protect\citeauthoryear{Vaswani \bgroup \em et al.\egroup
  }{2017}]{transformer}
Ashish Vaswani, Noam Shazeer, Niki Parmar, Jakob Uszkoreit, Llion Jones,
  Aidan~N. Gomez, Lukasz Kaiser, and Illia Polosukhin.
\newblock Attention is all you need.
\newblock In {\em NIPS}, pages 5998--6008, 2017.

\bibitem[\protect\citeauthoryear{Wang \bgroup \em et al.\egroup }{2016}]{ATAE}
Yequan Wang, Minlie Huang, Xiaoyan Zhu, and Li~Zhao.
\newblock Attention-based {LSTM} for aspect-level sentiment classification.
\newblock In {\em EMNLP}, pages 606--615, 2016.

\bibitem[\protect\citeauthoryear{Wang \bgroup \em et al.\egroup }{2020}]{RGAT}
Kai Wang, Weizhou Shen, Yunyi Yang, Xiaojun Quan, and Rui Wang.
\newblock Relational graph attention network for aspect-based sentiment
  analysis.
\newblock In {\em ACL}, pages 3229--3238, 2020.

\bibitem[\protect\citeauthoryear{Williams}{1992}]{Williams_1992}
Ronald~J. Williams.
\newblock Simple statistical gradient-following algorithms for connectionist
  reinforcement learning.
\newblock {\em Machine Learning}, 8(3-4):229--256, 1992.

\bibitem[\protect\citeauthoryear{Xing and Tsang}{2021}]{jair}
Bowen Xing and Ivor~W Tsang.
\newblock Out of context: A new clue for context modeling of aspect-based
  sentiment analysis.
\newblock {\em arXiv preprint arXiv:2106.10816}, 2021.

\bibitem[\protect\citeauthoryear{Xing and Tsang}{2022a}]{dignet}
Bowen Xing and Ivor Tsang.
\newblock Dignet: Digging clues from local-global interactive graph for
  aspect-level sentiment classification.
\newblock {\em arXiv preprint arXiv:2201.00989}, 2022.

\bibitem[\protect\citeauthoryear{Xing and Tsang}{2022b}]{kagrmn}
Bowen Xing and Ivor~W. Tsang.
\newblock Understand me, if you refer to aspect knowledge: Knowledge-aware
  gated recurrent memory network.
\newblock {\em IEEE Transactions on Emerging Topics in Computational
  Intelligence}, pages 1--11, 2022.

\bibitem[\protect\citeauthoryear{Xing \bgroup \em et al.\egroup
  }{2019}]{AA-LSTM}
Bowen Xing, Lejian Liao, Dandan Song, Jingang Wang, Fuzheng Zhang, Zhongyuan
  Wang, and Heyan Huang.
\newblock Earlier attention? aspect-aware {LSTM} for aspect-based sentiment
  analysis.
\newblock In {\em IJCAI}, pages 5313--5319, 2019.

\bibitem[\protect\citeauthoryear{Zhang and Qian}{2020}]{bigcn}
Mi~Zhang and Tieyun Qian.
\newblock Convolution over hierarchical syntactic and lexical graphs for aspect
  level sentiment analysis.
\newblock In {\em EMNLP}, pages 3540--3549, Online, 2020.

\bibitem[\protect\citeauthoryear{Zhang \bgroup \em et al.\egroup
  }{2019}]{asgcn}
Chen Zhang, Qiuchi Li, and Dawei Song.
\newblock Aspect-based sentiment classification with aspect-specific graph
  convolutional networks.
\newblock In {\em EMNLP}, pages 4568--4578, 2019.

\bibitem[\protect\citeauthoryear{Zhang \bgroup \em et al.\egroup
  }{2020}]{kgcap}
Bowen Zhang, Xutao Li, Xiaofei Xu, Ka-Cheong Leung, Zhiyao Chen, and Yunming
  Ye.
\newblock Knowledge guided capsule attention network for aspect-based sentiment
  analysis.
\newblock {\em IEEE ACM Trans. Audio Speech Lang. Process.}, 28:2538--2551,
  2020.

\end{thebibliography}
